\pdfoutput=1

\documentclass[11pt]{article}

\usepackage{ACL2023}

\usepackage{times}
\usepackage{latexsym}
\usepackage[T1]{fontenc}

\usepackage[utf8]{inputenc}

\usepackage{microtype}

\usepackage{inconsolata}

\usepackage{graphicx}
\usepackage[normalem]{ulem}
\useunder{\uline}{\ul}{}
\usepackage{multirow}
\usepackage{tabularx}
\usepackage{tcolorbox}
\usepackage{color}
\usepackage{caption}
\usepackage{float}
\usepackage{CJKutf8}

\let\OLDthebibliography\thebibliography
\renewcommand\thebibliography[1]{
  \OLDthebibliography{#1}
  \setlength{\parskip}{2pt} 
  \setlength{\itemsep}{2pt} 
}

\title{Towards Joint Modeling of Dialogue Response and Speech Synthesis\\based on Large Language Model}

\newcommand{\tb}[3]{\setlength{\tabcolsep}{#2mm}\begin{tabular}{#1}#3\end{tabular}}

\author{
\thanks{Xinyu Zhou and Delong Chen contributed equally.} 
\thanks{This work is partially supported by National Social Science Fund of China (20\&ZD295).}
\tb{@{}ccc@{}}{5}{
Xinyu Zhou  \begin{CJK*}{UTF8}{gbsn}(周欣宇)\end{CJK*}$^{1}$ & 
Delong Chen \begin{CJK*}{UTF8}{gbsn}(陈德龙)\end{CJK*}$^{2}$ & 
Yudong Chen \begin{CJK*}{UTF8}{gbsn}(陈玉东)\end{CJK*}$^{1}$ \\
}\\
\tb{cc}{5}{
$^{1}$Communication University of China \\  
$^{2}$Hong Kong University of Science and Technology \\
{\footnotesize \texttt{\{xinyuzhou, chenyd\}@cuc.edu.cn}, \ \ \texttt{delong.chen@connect.ust.hk}}
}}

\renewcommand\footnotemark{}

\begin{document}
\maketitle

\begin{abstract}
This paper explores the potential of constructing an AI spoken dialogue system that "\textit{thinks how to respond}" and "\textit{thinks how to speak}" simultaneously, which more closely aligns with the human speech production process compared to the current cascade pipeline of independent chatbot and Text-to-Speech (TTS) modules. We hypothesize that Large Language Models (LLMs) with billions of parameters possess significant speech understanding capabilities and can jointly model dialogue responses and linguistic features. We conduct two sets of experiments: 1) Prosodic structure prediction, a typical front-end task in TTS, demonstrating the speech understanding ability of LLMs, and 2) Further integrating dialogue response and a wide array of linguistic features using a unified encoding format. Our results indicate that the LLM-based approach is a promising direction for building unified spoken dialogue systems.\footnote{Codes and datasets are publicly available at \url{https://github.com/XinyuZhou2000/Spoken_Dialogue}.}
\end{abstract}

\section{Introduction}

\begin{figure}
    \centering
    \includegraphics[width=\linewidth]{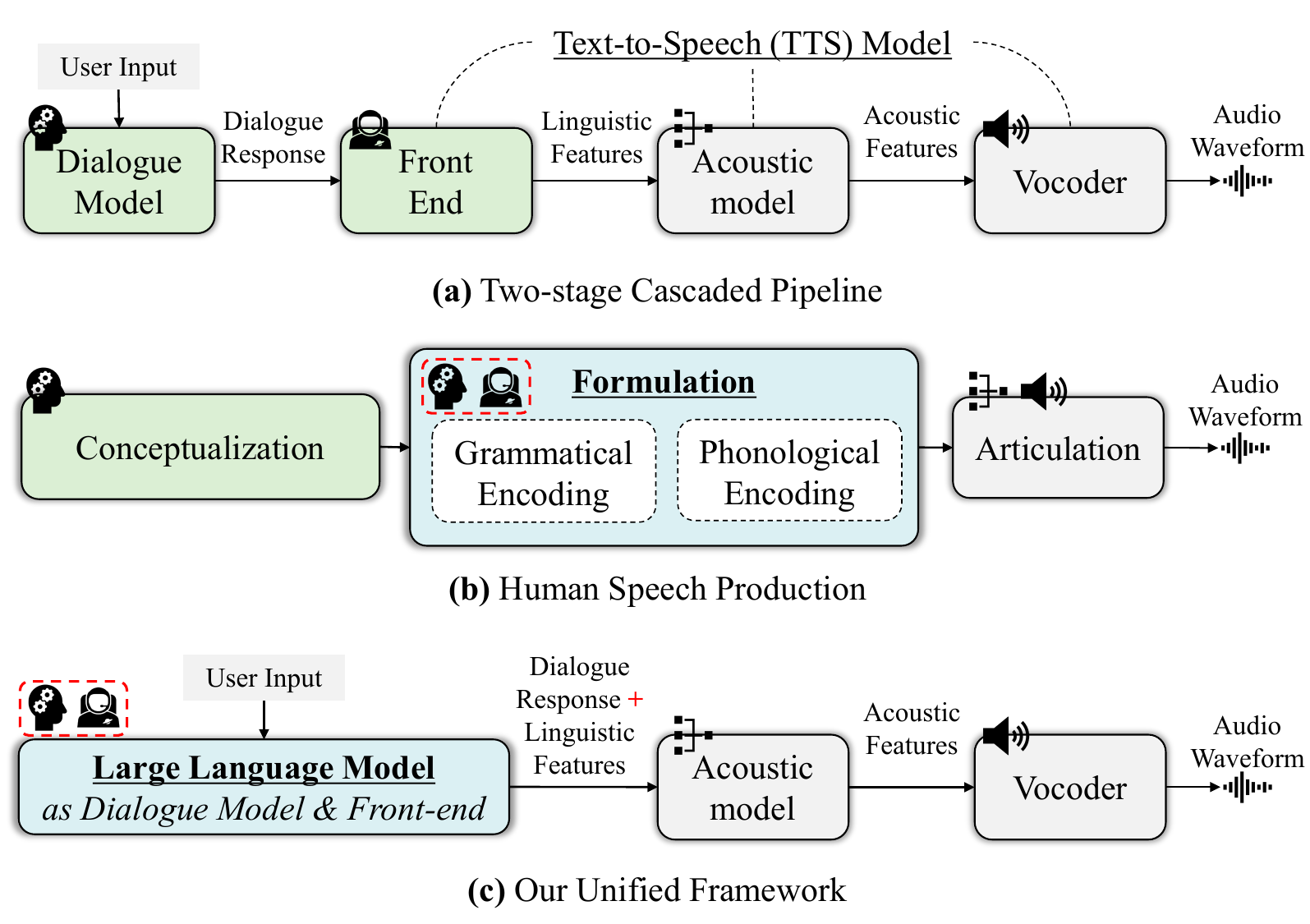}
    
    \caption{\footnotesize \textbf{A high-level comparison of different speech production processes. As noted by the \textcolor{red}{red dotted boxes}, the novel LLM-based unified framework proposed in this study can \textit{``think how to respond''} and \textit{``think how to speak''} at the same time, which aligns better with the speech production process of humans.}\vspace*{-5mm}}
    \label{fig:compare_tts_human}
\end{figure}

As we are developing more advanced AI systems, such as Large Language Model (LLM)-based chatbots like ChatGPT and GPT-4~\cite{OpenAI2023}, we also hope to establish natural, seamless, and efficient communication between humans and AI systems.  In addition to typing and reading through the screen, the speech channel represents a valuable alternative for interpersonal exchange, given its convenience and capacity to convey richer information than text alone. Recently, researchers from both academia and the industry have made successful attempts to concatenate AI chatbots with off-the-shelf text-to-speech (TTS)~\cite{Tan2021} modules as in Figure~\ref{fig:compare_tts_human} (a), representative applications include Siri, Xiaomi Xiaoai\footnote{\url{https://xiaoai.mi.com}} and Call Annie\footnote{\url{https://callannie.ai}}.

However, the expressivity and interactivity of speech responses synthesized by these two-stage cascaded models are heavily limited. The reasons are two-fold. \textit{\textbf{Firstly}}, TTS modules are usually based on small language models (\textit{e.g.,} BERT model with 0.1B parameters), which have limited capacity for understanding complex dialogue contexts. \textit{\textbf{Secondly}}, the dialogue response generation module (\textit{i.e.,} the LLM Chatbot) and the TTS module work independently. During speech synthesizing, the TTS module can not access the information from the dialogue context, which is proven to be valuable for generating plausible and appropriate speech responses.
  
The current two-stage pipeline also has a fundamental difference with our understandings of the human speech production process~\cite{Levelt1993}, where the \textit{``grammatical encoding''} and \textit{``phonological encoding''} are done in parallel within the ``conceptualization-formulation-articulation'' process, as shown in Figure~\ref{fig:compare_tts_human} (b). Inspired by this, we want to explore the possibility of building an AI speech dialogue system that \textit{``thinks how to respond''} and \textit{``thinks how to speak''} at the same time. In order to accomplish this goal, a model must possess a deep understanding of natural language and dialogue context, exhibit extensive world knowledge and commonsense, and demonstrate adequate learnability to handle text-speech joint modeling.

We hypothesize LLMs with (hundreds) billions of parameters (in comparison with BERT-based TTS front-ends~\cite{Chen2022} with only 0.3B parameters) are capable of achieving this goal. To verify this, in this paper, we provide two groups of experiments to demonstrate the possibility of building such LLM-based unified speech dialogue system. 

\textbf{\textit{Firstly}}, we get started with the prosodic structure prediction (Section~\ref{sec:prompt_prosodic_prediction}), a typical task within the TTS text analysis front-end, to showcase the speech understanding ability of LLMs. Results show that both prompting-based ChatGPT and fine-tuning based ChatGLM~\cite{Zeng2022} model achieve competitive performance against traditional methods. We also show that LLM can utilize linguistic knowledge to improve prediction accuracy.

\textbf{\textit{Secondly}}, we aim to further integrate a wide array of linguistic features into the model, and maintain LLM's dialogue capability at the same time (Section~\ref{sec:joint_learning}). To address the lack of a parallel dataset of dialogue response and linguistic annotations, we employ an automated dialogue context generation approach inspired by LongForm~\cite{Koeksal2023}, then train an LLM to produce both dialogue response speech features at the same time. Experiments show that LLM learns successfully.

\section{Related Work}

\subsection{Human Speech Production}
The process of human speech production is a long-standing research area. In 1993, Levelt~\cite{Levelt1993} proposed an encoding model for human speech production. First, concepts are generated, followed by the selection of appropriate vocabulary and the arrangement of these words according to grammatical rules. Then, the phonetics of the words are extracted in sequence, and motor programs are executed to initiate speech. The generation of spoken sentences is parallel and incremental, involving multiple stages of processing. Experiments~\cite{Schnur2011, Jaeger2012} prove that the phonetic planning of words begins as the grammatical structure of a sentence unfolds. Although there are many efforts to understand and explain human speech production process, TTS methods rarely take inspiration from these research results. To our best knowledge, this is the first study that attempts to build an AI system that imitates the simultaneous \textit{``grammatical encoding''} and \textit{``phonological encoding''} process of human speech production.

\subsection{TTS Front-end and Expressivity}
Typical TTS systems~\cite{Tan2021} usually consist of three main modules: front-end, acoustic model, and vocoder. The TTS Front-end models convert text into linguistic features, and are primarily BERT-based small language models, while the power of LLM is not well validated in this task yet. Hsu \textit{et al.}~\cite{Hsu2021} and Stephenson \textit{et al.}~\cite{Stephenson2022} have demonstrated that fine-tuning BERT can enhance the prosodic expression capabilities of TTS systems. Nevertheless, issues such as homograph ambiguity, ineffectiveness in stress, emotion and prosody still exist. Recent studies have explored the use of interactional resources~\cite{Chen2023}, such as breathing~\cite{Szekely2020}, laughter~\cite{Xin2023}, phonation type~\cite{Lameris2023}, filled pauses and prolongations~\cite{Li2023}, to improve the spontaneity and expressiveness. However, these studies have only focused on one single interactional resource, which limits their ability to capture rich and diverse subtle variations in natural conversation.

\subsection{LLMs for Speech Processing }
Understanding and generating speech signals are strongly related to natural language processing. With the recent explosion of LLM, many researchers in the field of speech processing also attempt to use LLMs to benefit speech or audio related tasks. AudioLM~\cite{Borsos2023} leverages a masked language model to capture the long-term structure and generate natural and coherent audio continuations given short prompts. SpeechGPT~\cite{Zhang2023}, a multi-modal large language model, leverages its inherent capabilities to perceive and generate multi-modal content. PromptTTS~\cite{Guo2023} and PromptTTS2~\cite{Leng2023} take prompts with both style and content descriptions as input to synthesize the corresponding speech. 

\section{Prosodic Structure Prediction based on Large Language Model}

\label{sec:prompt_prosodic_prediction}
\begin{figure}
    \centering
    \includegraphics[width=1\linewidth]{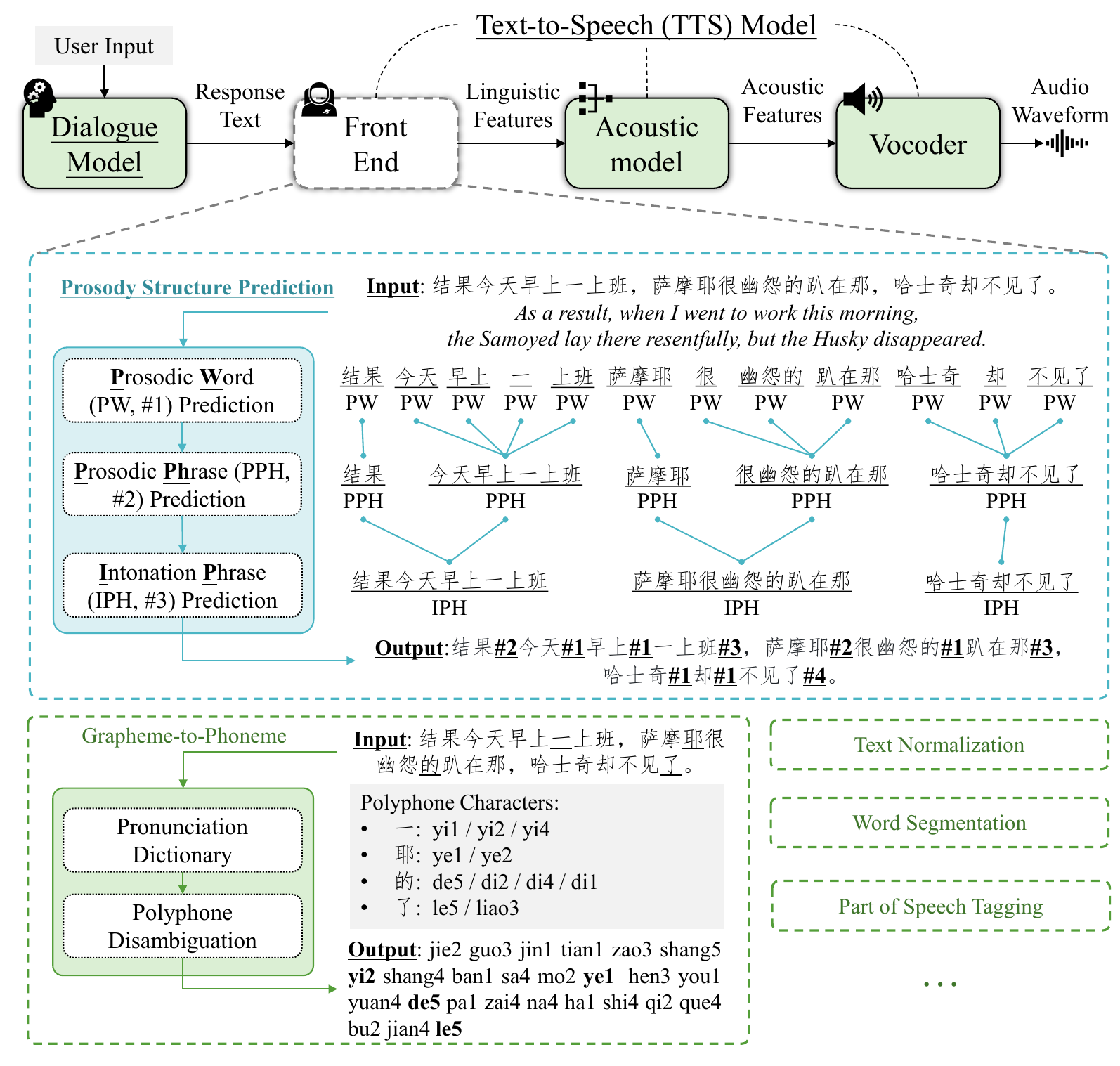}
    \caption{ \footnotesize \textbf{Standard pipeline of current spoken dialog systems.} A dialogue model generates a response to user input, and the TTS model (front-end→acoustic model→vocoder) converts text to audio subsequently.}\vspace*{-5mm}
    \label{fig:prosody_structure}
\end{figure}

Prosodic Structure Prediction (PSP) is a typical task in the Chinese TTS front-end~\cite{Chen2022}, among others like grapheme-to-phoneme prediction, text-normalization, word segmentation, part-of-speech tagging, etc. As illustrated in Figure~\ref{fig:prosody_structure}, a PSP model needs to identify multiple levels of prosody hierarchy, including Prosodic Word (PW), Prosodic Phrase (PPH), and Intonation Phrase (IPH), which can be denoted as \#1, \#2, and \#3 respectively in the output sentence. 

Prosodic structure is one of the most important linguistic features in Chinese TTS, and it is strongly related to the syntax of the sentence. In this section, we want to validate whether the LLMs, which have been well-proved to have superior semantic understanding abilities, but are trained on the text-only corpus, can handle this speech-related task. In the following, we present two methods for adapting LLM to the PSP task: prompting (Section~\ref{sec:prompting_llm_psp}), and fine-tuning (Section~\ref{sec:fine_tuning_llm_psp}).

\subsection{Prompting LLM for Prosodic Structure Prediction}
\label{sec:prompting_llm_psp}

\begin{figure}
    \centering
    \includegraphics[width=1\linewidth]{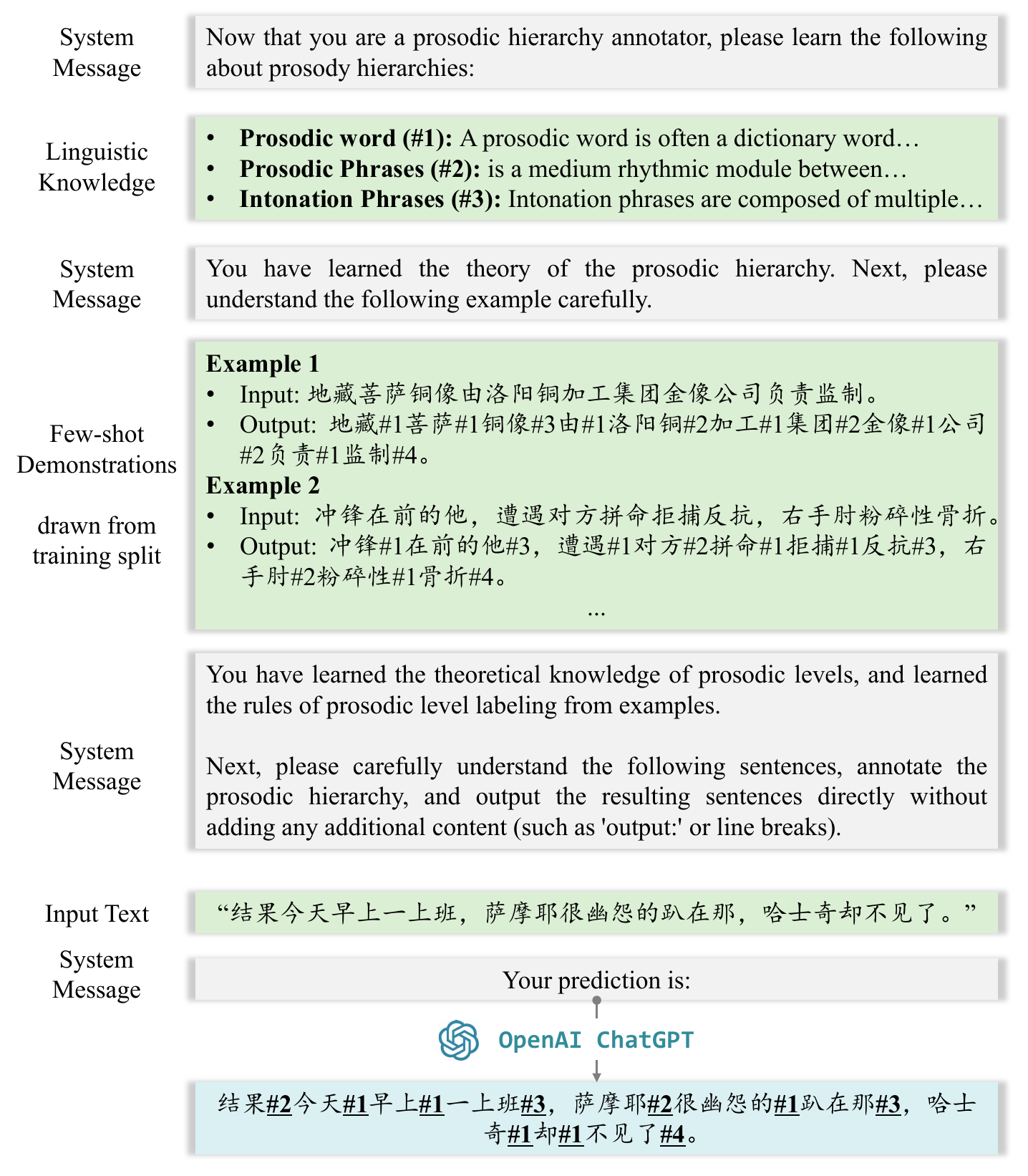}
    \caption{ \footnotesize \textbf{Our proposed prompt structure for LLM (ChatGPT)-based prosodic structure prediction.} We incorporate expert linguistic knowledge and few-shot demonstrations to enable LLMs to perform the prosodic structure prediction task.}\vspace*{-5mm}
    \label{fig:prompt_prosody_prediction}
\end{figure}

Prompting is the most convenient way to adapt an instruction-following LLM to new tasks. In Figure~\ref{fig:prompt_prosody_prediction}, we present an overview of our proposed prompt structure for PSP on LLM, which consists of linguistic knowledge of Chinese prosodic structure, few-shot demonstrations as in-context learning examples, input sentence, and interleaved system messages to explain each part to the LLM.

\textbf{\textit{Linguistic Knowledge}} contains formal definitions of Chinese prosodic structure summarized from recognized research literature~\cite{Cao2003}. It describes distinct characteristics and positions within sentences and phrases of three levels of prosodic structure in Chinese. 

\textbf{\textit{Few-shot Demonstration}} provides input-output pairs to LLM for in-context learning. Examples are either randomly drawn from the training split or carefully selected based on the assessment of their representativity and quality from the linguistic perspective. The maximum number of few-shot demonstrations is 16, as more examples would exceed the context window length of LLM.

\subsection{Fine-tuning LLM for Prosodic Structure Prediction}
\label{sec:fine_tuning_llm_psp}

Context window length is a crucial limit for prompting-based methods, as it prohibits the LLM from learning from more (than 16) training examples. Furthermore, all model parameters remain fixed and unlearnable, resulting in limited learning capacity. To address these constraints, we propose the fine-tuning of a Large Language Model (LLM) to enhance Prosodic Structure Prediction learning from a substantially larger number of training examples, up to 8,000.

It has been proved that using a Pretrained Language Model (PLM) such as BERT~\cite{Devlin2018} to be the initialization of the PSP model is beneficial, such as SpanPSP~\cite{Chen2022}, J-TranPSP~\cite{Shen2022}, and MLC-PSP~\cite{Chen2023a}, our methodology of fine-tuning LLM has some difference from them. Despite the difference in model scale (0.1B vs 6B), previous BERT-based methods regard PSP as a \textit{token classification} problem, where the model needs to determine whether there is a prosodic boundary  after each character and what level is it. In contrast, here we formalize PSP as a sequence-to-sequence (Seq2seq) prediction task, where input $x$ is the raw sentence and the output $y$ is a string of character sequence with ``\#$n$'' ($n\in\{1,2,3\}$) notation of prosodic structure.

We apply standard cross-entropy loss for auto-regressive language modeling as the learning objective, and we only calculate the loss on output tokens. We add a prefix $c$ of ``Please perform prosodic prediction on the given sentence:'' into the input for better initialization. The following is the loss function $\mathcal{L}(\theta)$ of the LLM $\theta$, where $N$ is the number of training samples: $\mathcal{L}(\theta) = -\sum_{i=1}^{N} \log p_{\theta}(y_{i} | x_{i}, c_{i})$.


\subsection{Experiment Setup}

\textbf{Dataset}. We utilize the DataBaker open-source Chinese Standard Mandarin Speech Corpus\footnote{\url{https://www.data-baker.com/data/index/TNtts}}, which contains 10-hour speech recordings of 10,000 sentences with an average length of around 16 words per sentence. It was articulated by a single Chinese female speaker. The corpus also encompasses diverse domains, including news, novels, technology, entertainment, etc. 

Furthermore, the dataset is enriched with various linguistic annotations, including character, pinyin, and prosodic hierarchy information, as well as phoneme level interval and boundary data. Annotations for prosodic hierarchy comprise PW (\#1), PPH(\#2), IPH (\#3), and the end of a sentence (\#4). We discard the \#4 annotations as every sample is a single sentence and only has a ``\#4'' in the end. The remaining labels collectively form a hierarchical prosodic tree structure with three distinct layers.

We split 10k samples with an 8:1:1 ratio for training, validation, and testing. Few-shot demonstrations are drawn from the 8k training split. For the ``random'' selection setting, we sample demonstrations randomly three times and report the averaged performance.

\textbf{Implementation Details}. For the prompting-based method, we test the OpenAI's \texttt{text-davinci-002} API (ChatGPT) and the ChatGLM2-6B model. For the fine-tuning-based method, we only unitize the ChatGLM2-6B model due to the limitation of computational resources. We apply P-tuning-v2~\cite{Liu2022} for parameter-efficient fine-tuning using the official codebase~\footnote{\url{https://github.com/THUDM/ChatGLM2-6B/tree/main/ptuning}}. We used a single NVIDIA A100 GPU for both training and testing.

\subsection{Ablation Study}

\begin{figure}
    \centering
    \includegraphics[width=1\linewidth]{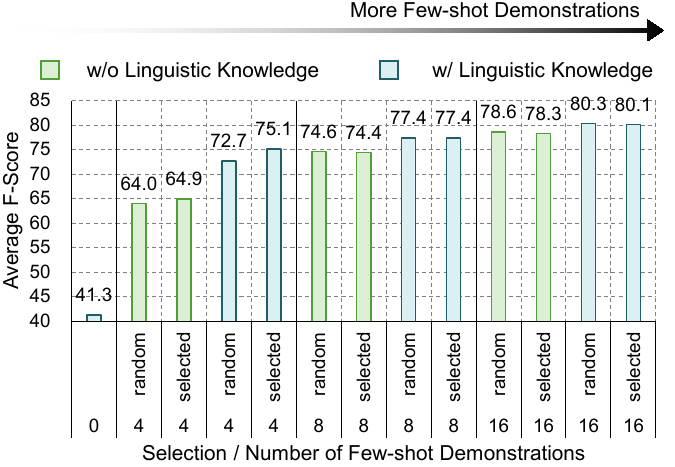}
    \caption{ \footnotesize \textbf{Ablation study of prompting ChatGPT based PSP}. We compared different numbers of few-shot demonstrations, selection of few-shot demonstrations, and variants of with (w/) or without (w/o) linguistic knowledge. }\vspace*{-5mm}
    \label{fig:few_shot_ablations}
\end{figure}

\begin{table*}[]
\centering
\caption{ \footnotesize \textbf{Benchmarking of PSP Models}. We compared the F-Score of the traditional BERT-based method SpanPSP and our newly proposed LLM-based methods (prompting or fine-tuning) using two LLMs with different scales (ChatGPT and ChatGLM).}
\label{tab:benchmarking-psp}
\resizebox{0.8\textwidth}{!}{%
\begin{tabular}{cccccc}
\hline\hline
\textbf{Model (\#Parameters)} &
  \textbf{Variation} &
  \textbf{PW   \#1} &
  \textbf{PPH   \#2} &
  \textbf{IPH   \#3 }&
  \textbf{Average} \\ \hline
\multirow{2}{*}{SpanPSP (0.1B)}    & Databaker   Pretrained             & {\ul \textbf{96.35}} & 69.34                & 65.64                & 77.11                \\
                            & PeopleDaily   Pretrained           & 89.20                & 71.08                & 79.12                & 79.80                \\ \hline
\multirow{3}{*}{ChatGPT (175B)}    & Knowledge   Only                   & 61.87                & 27.27                & 34.78                & 41.31                \\
                            & 16   Random Examples               & 88.51                & 69.40                & 77.91                & 78.61                \\
                            & Knowledge   + 16 Selected Examples & 90.12                & 69.40                & {\ul \textbf{80.85}} & 80.12                \\ \hline
\multirow{2}{*}{ChatGLM2-6B} & Knowledge   + 16 Selected Examples & N/A                  & N/A                  & N/A                  & N/A                  \\
                            & Fine-tuned                         & 93.86                & {\ul \textbf{73.28}} & 80.00                & {\ul \textbf{82.38}} \\ \hline
\hline
\end{tabular}%
}
\end{table*}

We first provide ablations for the prompting-based approach. Following previous works on PSP tasks, we use F-Score as the evaluation metric. As it can be seen from Figure~\ref{fig:few_shot_ablations}, the number of few-shot demonstrations makes a significant impact. Four in-context examples lead to +22.7\% improvements to zero-shot setting (41.3\%$\rightarrow$64.0\%), while incorporating linguistic knowledge brings another around +8.7\% improvements ($\rightarrow$72.7\%), and further, when swapping random demonstrations to carefully selected high-quality demonstrations, we receive another +2.4\% performance gain ($\rightarrow$75.1\%).

\begin{table}[]
\centering
\caption{ \footnotesize \textbf{Ablations of removing each level of linguistic knowledge.} Expert knowledge is especially useful for higher levels of prosodic structure prediction (\textit{i.e.,} PPH and IPH).}
\label{tab:knowledge-level-ablation}
\resizebox{0.95\linewidth}{!}{%
\begin{tabular}{ccccc}
\hline\hline
\textbf{\begin{tabular}[c]{@{}c@{}}Knowledge\\      Ablation\end{tabular}} &
  \textbf{\begin{tabular}[c]{@{}c@{}}PW   \#1\\      F-Score\end{tabular}} &
  \textbf{\begin{tabular}[c]{@{}c@{}}PPH   \#2\\      F-Score\end{tabular}} &
  \textbf{\begin{tabular}[c]{@{}c@{}}IPH   \#3\\      F-Score\end{tabular}} &
  \textbf{\begin{tabular}[c]{@{}c@{}}Average\\      F-Score\end{tabular}} \\ \hline
w/o   \#1       & {\ul \textbf{88.54}} & 64.66                & 78.30                & 77.17                \\ \hline
w/o   \#2       & 87.57                & {\ul \textbf{61.63}} & 79.09                & 76.10                \\ \hline
w/o   \#3       & 87.72                & 64.69                & {\ul \textbf{78.14}} & 76.85                \\ \hline
Default   (all) & {\ul \textbf{88.14}} & {\ul \textbf{65.03}} & {\ul \textbf{79.52}} & {\ul \textbf{77.56}} \\ \hline
\hline
\end{tabular}%
}
\end{table}

We further ablate different levels of linguistic knowledge in Table~\ref{tab:knowledge-level-ablation}. It shows that linguistic expert knowledge plays a crucial role in the prediction of Prosodic Phrase (\#2) and Intonational Phrase (\#3). We hypothesize it is caused by different difficulties of \#1 to \#3 predictions --  \#1 usually appears at word boundaries, while identifying \#2 and \#3 is not that straightforward.

\subsection{Benchmarking LLM-based PSP}

\textbf{Baseline.} SpanPSP~\cite{Chen2022} is a classical character-level BERT-based model for the PSP task, which is based on a relatively small language model \texttt{bert-base-chinese}\footnote{\url{https://huggingface.co/bert-base-chinese}} with only 0.1B parameters. We use their official checkpoints and codebase\footnote{\url{https://github.com/thuhcsi/SpanPSP}} for evaluation. 

We provide benchmarking results in Table~\ref{tab:benchmarking-psp}. It reveals that carefully crafted linguistic knowledge and selected examples (\textit{i.e.,} ``Knowledge + 16 Selected Examples'' variation) enable ChatGPT to outperform the traditional method SpanPSP (80.12\% vs. 79.80\%), but such a prompting-based learning strategy failed (N/A) at smaller open-source LLM (ChatGLM) due to its limited instruction-following ability. However, it shows that fine-tuning smaller LLM can outperform prompting larger LLM (82.38\% vs. 80.12\%), as it can access more training samples (8k training set vs. the maximum of 16 in-context examples).

\begin{figure*}[h!]
    \centering
    \includegraphics[width=1\linewidth]{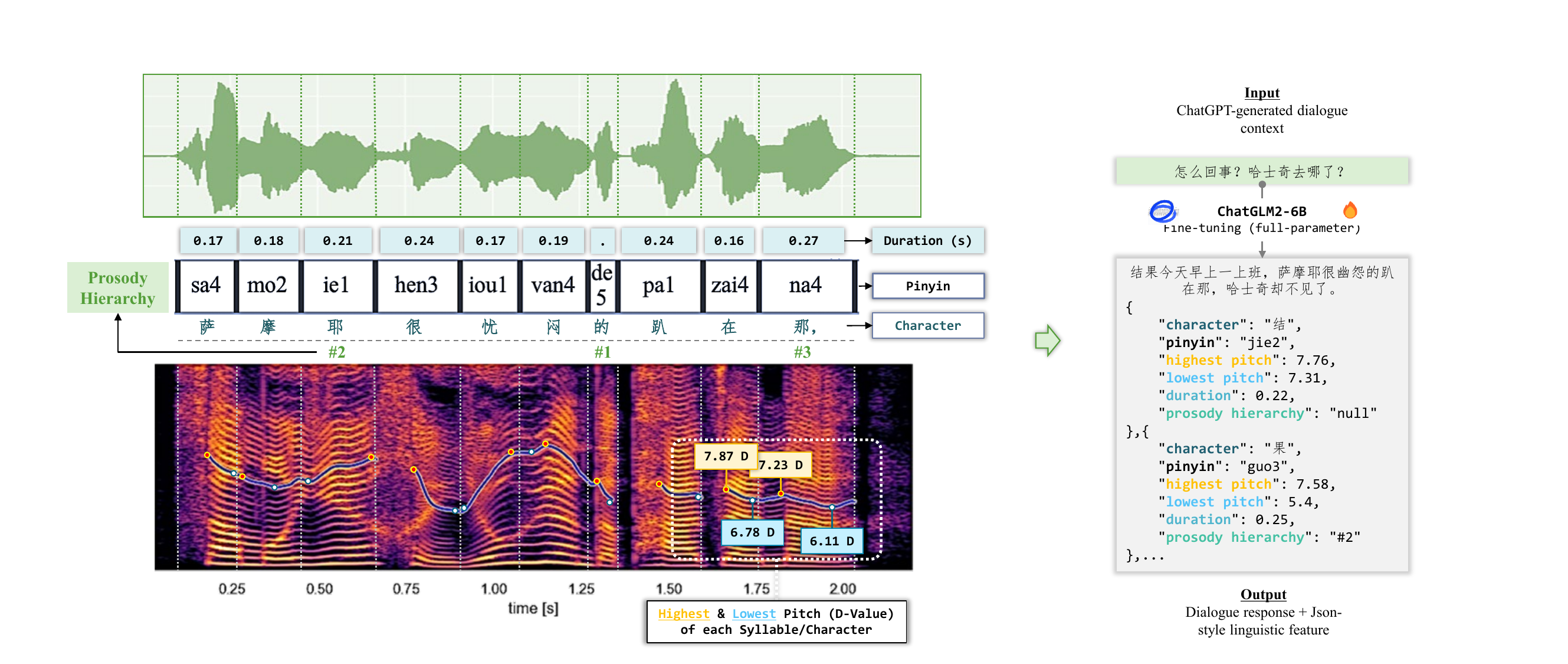}
    \caption{ \footnotesize \textbf{Left:} an overview of linguistic feature extraction. We automatically extract a wide array of linguistic features, including character, duration, pinyin, prosody hierarchy, highest pitch and lowest pitch ($\mathcal{D}$-Value). \textbf{Right:} the illustration of data formatting. We encode extracted linguistic features into JSON-formatted strings, such that they can be fed to LLM directly as learning targets.}
    \label{fig:visualize_linguistic_features}\vspace*{-5mm}
\end{figure*}

\section{Joint Prediction of Dialogue Response and Linguistic Features}
\label{sec:joint_learning}

In the last section, we have shown some positive results proving LLMs are competitive at a typical front-end task in Chinese TTS. Here in this section, we want to go beyond just a single task in TTS, and validate the possibility of building a LLM-based system that can handle versatile tasks in Chinese TTS front-end (Figure~\ref{fig:prosody_structure}), and also maintain its dialogue capability of generating coherent responses to user queries simultaneously. By implementing such a model that jointly predicts dialogue response (\textit{i.e., ``think how to respond''}) and linguistic features (\textit{i.e., ``think how to speak''}) at the same time, we could have an AI system that communicates with humans in a way that is more similar to the human speech production process~\cite{Levelt1993}, where the ``grammatical encoding'' and ``phonological encoding'' are done in parallel.

\subsection{Methodology}

\textbf{Dialogue Context Generation}. Our objective is to investigate the feasibility of constructing a unified model capable of simultaneously generating coherent responses to user queries in dialogues and diverse fine-grained linguistic features for TTS. Unfortunately, the DataBaker dataset only comprises isolated sentence recordings, and there are not any other datasets having dialogue context and parallel speech recordings or annotations. Drawing inspiration from the LongForm approach~\cite{Koeksal2023}, we prompt ChatGPT to anticipate the dialogue context and transform it into a dataset of single-turn dialogues:

\begin{table}[h!]
\centering
\begin{minipage}{\linewidth}\vspace{0mm}
\centering
\footnotesize
\begin{tcolorbox}[colback=green!0.7!white,colframe=green!30!black,title=\textbf{Prompt for Dialogue Context Generation}]
\centering
\begin{tabular}{p{\columnwidth} c}

\textcolor{blue}{\textbf{\texttt{\#\#\# System Message:}}}\\
Please generate the most likely sentence spoken by A based on B's response. & \\


\textcolor{blue}{\textbf{\texttt{\#\#\# User:}}}\\
A: \\
B: \textit{``When I went to work this morning, the Samoyed lay there resentfully, but the Husky disappeared.''} \\

\hrulefill & \\

\textcolor{blue}{\textbf{\texttt{\#\#\# ChatGPT:}}}\\
A: \textit{``What's going on? Where did the Husky go?''}
\end{tabular}
\end{tcolorbox}
\end{minipage}
\end{table}

\textbf{Linguistic Feature Extraction}. As shown in Figure~\ref{fig:visualize_linguistic_features} left, we automatically extract the following four categories of linguistic attributes: characters, their corresponding duration, pinyin (phonetic transcription representing character pronunciation), prosodic hierarchy, and the highest and lowest pitch values ($\mathcal{D}$-Value). The use of $\mathcal{D}$-value is inspired by Shen Jiong's theory~\cite{Shen1985}: the D-value is a logarithmic scale used to describe pitch and quantifies the relationship between a pitch ($F$) in Hertz (Hz) and a reference frequency ($F_0$). It provides a measure of pitch variation, which is especially useful for observing pitch contours in speech. The formal definition of ($\mathcal{D}$-Value) is: $\mathcal{D} = 5 \times {\log}_2 (F/F_0)$.


\textbf{Data Formatting}. As shown in the right side of Figure~\ref{fig:visualize_linguistic_features}, we format extracted linguistic features into a string of JSON-style dictionaries, and concatenate it with the response text generated by ChatGPT, together serving as the learning target. Such implementation realizes joint learning in a seamless way and enjoys simplicity over the traditional method, where different types of outputs (response, various linguistic features) are usually produced by different models, or one model with different task-specific heads~\cite{Bai2021}. Our approach also shares some similarities with recent advances in LLM research, such as RT-2~\cite{Brohan2023} from Google DeepMind, where the LLM are trained to produce not only natural language output but also some continuous values.

\subsection{Experiment Setup}

\textbf{Training}. Empirically, we found that P-tuning (as used in fine-tuning-based PSP in Section~\ref{sec:fine_tuning_llm_psp}) failed to learn how to generate dialogue response and JSON-style linguistic features. Therefore, for this section, we turn to use full-parameter fine-tuning to enable more learning capacity. We use 4-bit quantization to boost memory efficiency, as JSON-style encoding takes much longer context than that in the PSP task (maximum 1.6k tokens vs. 128 tokens).

\textbf{Testing}. Based on our data formatting (Figure~\ref{fig:visualize_linguistic_features}), given a user utterance as input, the model will first give its dialogue response, then the JSON-style linguistic feature of each word in the response sentence subsequently. However, this poses a challenge for the evaluation of the linguistic feature, since for unseen testing quires, the LLM-outputted response would be different from the ground truth response, thus making them not comparable. To solve this issue, we use the ground truth response as a generation prefix, and then try to parse the generated dictionary and compare them with ground truth linguistic features.

\subsection{Experiment Results}

\begin{table}[]
\centering
\caption{ \footnotesize \textbf{Evaluation result of LLM produced linguistic features.} Most of the output JSON-style is incorrect grammar and parsable, and the majority of these parsable characters can be matched with ground truth. However, we can observe a notable train-test performance gap, meaning that the model suffers from overfitting.}
\label{tab:joint-json-result}
\resizebox{\linewidth}{!}{%
\begin{tabular}{ccccc}
\hline\hline
 &
  \textbf{\begin{tabular}[c]{@{}c@{}}Parsable\\      Samples\end{tabular}} &
  \textbf{\begin{tabular}[c]{@{}c@{}}Matched\\      Characters\end{tabular}} &
  \textbf{\begin{tabular}[c]{@{}c@{}}Matched\\      Pinyin\end{tabular}} &
  \textbf{\begin{tabular}[c]{@{}c@{}}Matched\\      Prosody\end{tabular}} \\ \hline
Training Split &
  95.90\% &
  86.88\% &
  98.79\% &
  97.75\% \\
Testing Split &
  89.70\% &
  69.26\% &
  86.29\% &
  77.70\% \\ \hline\hline
\end{tabular}%
}
\end{table}

\begin{figure*}
    \centering
    \includegraphics[width=1\linewidth]{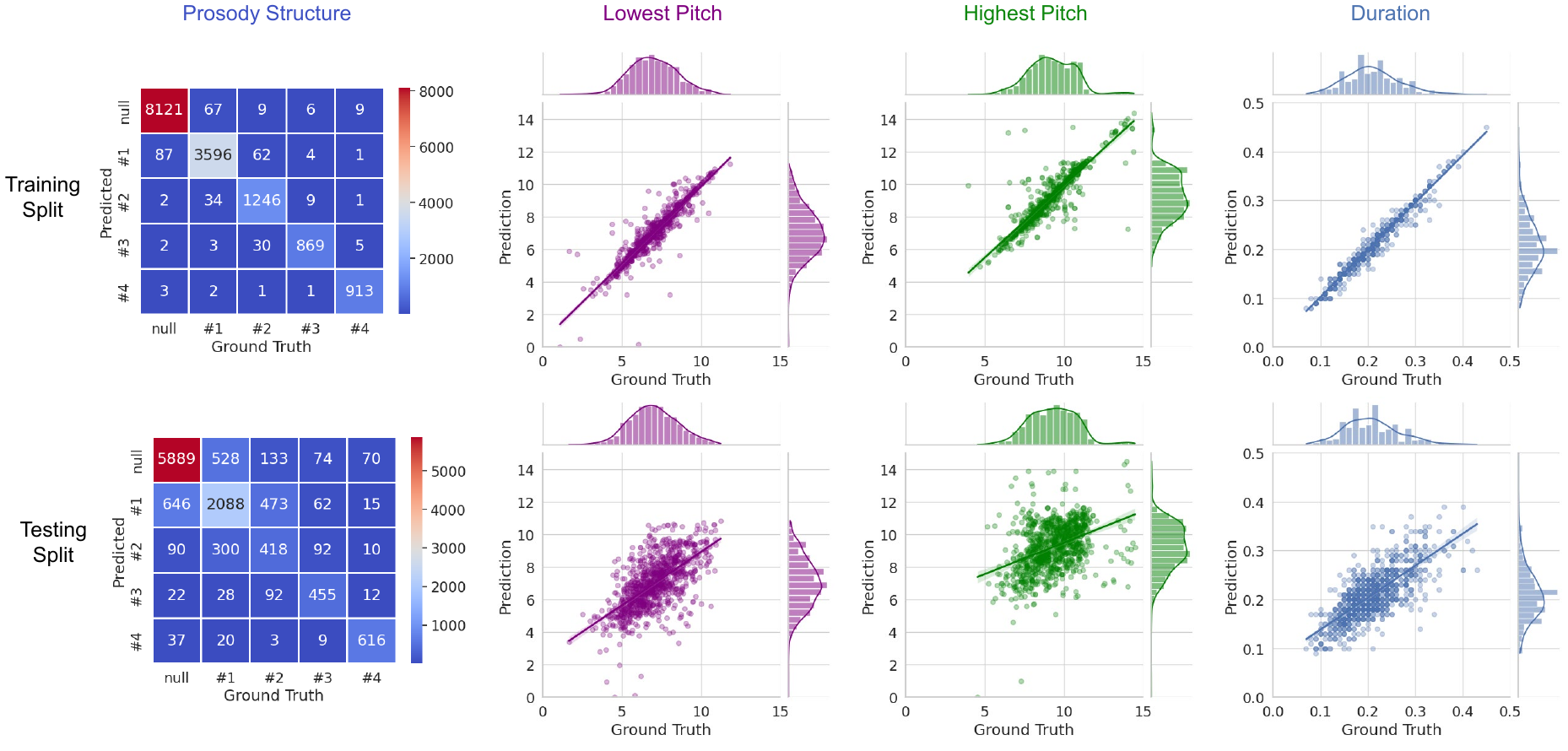}
    \caption{ \footnotesize \textbf{Evaluations results of fine-tuning ChatGLM2-6B on joint dialogue response and linguistic features.} Visualizations show that the model fits the training data quite well, showing the feasibility of our proposed joint learning approach. But possibly due to insufficient dataset scale, the generalization ability of the model is somewhat weak.}\vspace*{-5mm}
    \label{fig:fine_tune_visualizations}
\end{figure*}

In Table~\ref{tab:joint-json-result}, we provide the evaluation result of 1k testing samples and randomly sampled 1k training samples. As can be seen, the model performed quite well on the training set, achieving 95.90\% parsable samples, 86.88\% matched characters, and 98.79\% matched Pinyin, showing that it successfully fit the JSON-format data. When tested on unseen samples, the model successfully generated Json-style linguistic features with an 89.70\% success rate. However, due to the limited capacity of the small LLM, missing characters were frequently observed, resulting in only 69.26\% of characters from the ground truth being found in the generated results. Within those matched characters, the model achieved an 86.29\% success rate in producing matched Pinyin, and a 77.70\% success rate of matched prosody structure annotation. 

In Figure~\ref{fig:fine_tune_visualizations}, we visualize the model predictions versus the ground truth of continuous values. Again, we observed that the model effectively fit the training set and demonstrated a certain level of generalization ability when applied to new data.

\section{Discussion}

In this study, we presented two groups of experiments to validate the possibility of building an LLM-based spoken dialog system that \textit{``thinks how to respond''} and \textit{``thinks how to speak''} at the same time. In the first group of experiments, we proved that LLM is a competitive prosodic structure predictor, which means that its rich world knowledge and semantic understanding ability acquired from text-only pretraining can transferred to benefit speech-related tasks. Based on this observation, we further involve many other linguistic features in our second group of experiments and further proved that it is possible for LLM to learn to generate dialogue response and speech features at the same time. However, there are still several noticeable limitations of this study, which are summarized from the following three perspectives:

\textbf{Model Perspective}. The training cost of LLM is high. Additionally, the auto-regressive decoding of Json-style linguistic features is quite time-consuming -- processing a single sentence and its linguistic features takes at least 15 seconds (for long sentences, it could be 40+ seconds).

\textbf{Data Perspective}. The current training dataset consists of only 8k samples, which is insufficient and has led to a substantial over-fitting phenomenon. Speech style in the dataset is limited, as it was sourced from a single speaker, primarily containing formal read recordings, lacking the nuances inherent in natural conversations.

\textbf{Expressivity Perspective}.  According to interactional linguistics studies~\cite{CouperKuhlen2017}, finer-grained annotation system by comprehensively and meticulously annotating the collected speech dataset with interactional resources like voice quality, phonation type, breath patterns, repair, interjection, pause, prolongation, etc, will further increase the expressivity.

\textbf{System Perspective}. It is important to note that so far this study does not include subsequent acoustic models and vocoders and is not able to generate audio waveform, we only use speech linguistic features to represent speech information.

\bibliographystyle{acl_natbib}
{\footnotesize
\bibliography{anthology.bib}}

\end{document}